\pdfoutput=1
\documentclass[11pt]{article}

\usepackage[]{emnlp2021}

\usepackage{times}
\usepackage{latexsym}

\usepackage[T1]{fontenc}
\usepackage[utf8]{inputenc}

\usepackage{microtype}
\usepackage{times}
\usepackage{latexsym}
\usepackage{amsfonts}
\usepackage{array}
\usepackage{amsmath}
\usepackage{mathtools}
\usepackage{booktabs}
\usepackage{algpseudocode}
\usepackage[ruled,vlined]{algorithm2e}
\usepackage{amssymb}
\usepackage{url}
\usepackage{color}
\usepackage{multirow}
\usepackage{pifont}
\usepackage{graphicx}
\usepackage{subfigure}
\usepackage{lipsum}
\usepackage{footmisc} 
\usepackage{footnote}
\usepackage{tablefootnote}
\usepackage{afterpage}
\usepackage{adjustbox}

\newcommand{\cmark}{\ding{51}}%
\newcommand{\xmark}{\ding{55}}%
\graphicspath{{figures/}}
\newcommand*{\affaddr}[1]{#1} % No op here. Customize it for different styles.
\newcommand*{\affmark}[1][*]{\textsuperscript{#1}}
\newcommand*{\email}[1]{\texttt{#1}}
\usepackage{xstring}

\title{Text AutoAugment: \\ 
Learning Compositional Augmentation Policy for Text Classification}
\author{Shuhuai Ren\affmark[1], Jinchao Zhang\affmark[3], Lei Li\affmark[1], Xu Sun\affmark[1,2], Jie Zhou\affmark[3]\\
\affaddr{\affmark[1]MOE Key Lab of Computational Linguistics, School of EECS, Peking University}\\
\affaddr{\affmark[2]Center for Data Science, Peking University}\\
\affaddr{\affmark[3]Pattern Recognition Center, WeChat AI, Tencent Inc, China}\\
\email{\{shuhuai\_ren, lilei\}@stu.pku.edu.cn, xusun@pku.edu.cn}\\
\email{\{dayerzhang, jiezhou\}@tencent.com}\\
}

\begin{document}
\maketitle
\begin{abstract}
Data augmentation aims to enrich training samples for alleviating the overfitting issue in low-resource or class-imbalanced situations.
Traditional methods first devise task-specific operations such as Synonym Substitute, then preset the corresponding parameters such as the substitution rate artificially, which require a lot of prior knowledge and are prone to fall into the sub-optimum. 
Besides, the number of editing operations is limited in the previous methods, which decreases the diversity of the augmented data and thus restricts the performance gain.
To overcome the above limitations, we propose a framework named \textbf{T}ext \textbf{A}uto\textbf{A}ugment (TAA) to establish a \emph{compositional} and \emph{learnable} paradigm for data augmentation. 
We regard a combination of various operations as an augmentation policy and utilize an efficient Bayesian Optimization algorithm to automatically search for the best policy, which substantially improves the generalization capability of models. 
Experiments on six benchmark datasets show that TAA boosts classification accuracy in low-resource and class-imbalanced regimes by an average of 8.8\% and 9.7\%, respectively, outperforming strong baselines.\footnote{Our code is available at \url{https://github.com/lancopku/text-autoaugment}}
\end{abstract}

\section{Introduction}
\label{sec:Intro}

\begin{figure}[t!]
    \centering
    \includegraphics[width=0.45\textwidth]{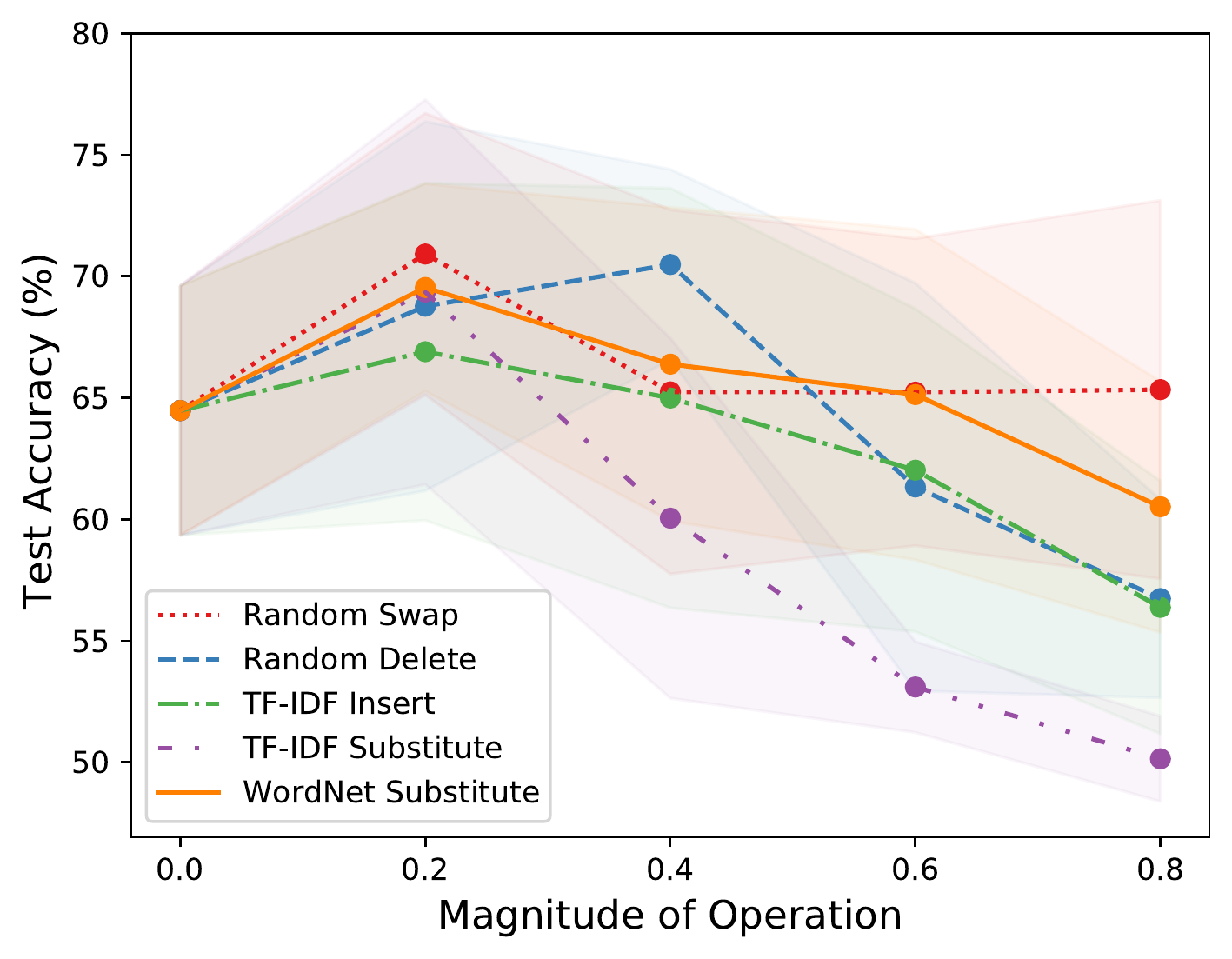}
    \caption{Test accuracy on IMDB dataset with different editing operations and parameters for data augmentation. The operations and parameters without elaborate selection lead to a degrade of performance.}%Here $n_\text{aug}=8$ for all augment operations.%}
    \label{Fig:imdb_m}
\end{figure}

\begin{figure*}[t!]
    \centering
    \includegraphics[width=\textwidth]{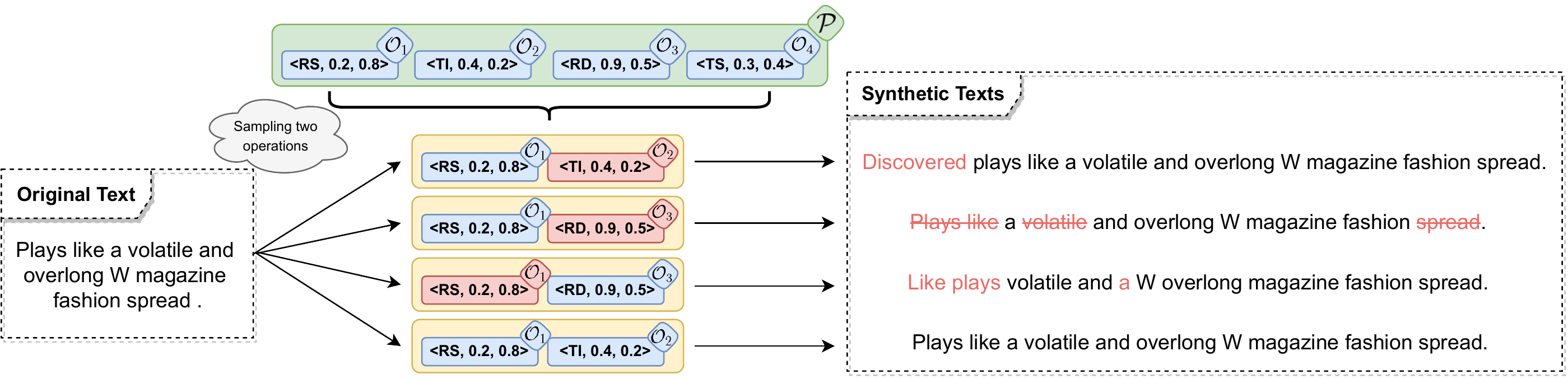}
    \caption{The structure and usage of augmentation policy $\mathcal{P}$. A policy $\mathcal{P}$ consists of $N$ atomic editing operations $\mathcal{O}=\left<t,p,\lambda\right>$. Given a text, we randomly pick $N^*$ operations from $\mathcal{P}$ then apply the operations sequentially. In this case, $N=4$ and $N^*=2$. {\color[HTML]{EA6B66} Red operation} $\mathcal{O}$ indicates it is finally applied to the text according to probability $p$. The modification to the original text cased by it is also showed in {\color[HTML]{EA6B66} Red}. We use {\bf RS} (Random Swap), {\bf RD} (Random Delete), {\bf TI} (TF-IDF Insert), {\bf TS} (TF-IDF Substitute) for abbreviation. Best viewed in color.}
    \label{Fig:augmentation policy}
\end{figure*}

Model performance on Natural Language Processing (NLP) tasks, such as text classification, often heavily depends on the size and the quality of the training data. 
However, it is time-consuming and labor-intensive to obtain sufficient training instances, and models face low-resource regimes most of the time. 
Data augmentation~\cite{Transformation-Invariance, Going-deeper-with-convolutions, eda} aims to enlarge the training dataset by synthesizing additional distinct and label-invariant instances based on raw instances to improve the model performance. 
Data augmentation approaches demonstrate their superiority in many scenarios, especially when data resources are insufficient~\cite{learning-data-manipulation, not-enough-data, da4limited-data-learning} and class-imbalanced~\cite{learning-to-reweight, learning-data-manipulation}. 

% Previous data augmentation methods in the NLP area can be roughly divided into two categories: generation-based and editing-based methods. 
% Back-translation~\cite{improving-MT, enhancement-MT} and conditional generation~\cite{contextual-augmentation, not-enough-data} are two kinds of popular generation-based methods, which have the advantage on instance fluency and label preservation but suffer from the model pre-training/training and decoding cost. 
Previous data augmentation methods can be roughly divided into two categories: generation-based~\cite{improving-MT, enhancement-MT, contextual-augmentation, not-enough-data, da4al} and editing-based methods~\cite{eda, uda}. 
Generation-based methods utilize conditional generation models~\cite{seq2seq, ott2019fairseq} to synthesize the paraphrases~\cite{Submodular-Paraphrasing, PARABANK} of the original sentences, which have advantages in instance fluency and label preservation but suffer from the heavy cost of model pre-training and decoding. 
Editing-based methods instead apply label-invariant sentence editing operations (swap, delete, etc.) on the raw instance, % to create a synthetic one, 
which are simpler and more efficient in practice. 
However, the editing-based methods are sensitive to the preset hyper-parameters including 
the type of the applied operations 
% the type of editing operations to be applied on the sentence 
and the proportion of words to be edited. 
% that determine which editing operations to be applied on one sentence and how many words to be edited.
As shown in Figure~\ref{Fig:imdb_m}, we probe the classification accuracy on IMDB dataset~\cite{IMDB} with different editing operations and magnitudes~(proportion of edited words). We find that inappropriate hyper-parameter settings, e.g, TF-IDF Substitute~\cite{uda} with a magnitude larger than 0.4, lead to inferior results. Therefore, heuristic hyper-parameter setting is prone to fall into the sub-optimum and lacks effectiveness. 
Besides, most of the editing-based methods~\cite{contextual-augmentation, learning-data-manipulation, improving-MT, enhancement-MT, good-enough-compositional} only apply a single operation on the sentence once a time, which restrict the diversity of the augmented dataset and thus limit the performance gain.

To overcome the limitations above, we propose a framework named \textbf{T}ext \textbf{A}uto\textbf{A}ugment (TAA) to establish a \emph{learnable} and \emph{compositional} paradigm for the data augmentation. 
Our goal is to automatically learn the optimal editing-based data augmentation policy for obtaining a higher quality augmented dataset and thus enhancing the target text classification model. 
We design an augmentation policy as a set of various editing operations:  $\operatorname{policy} = \{\operatorname{op}_1, \cdots, \operatorname{op}_N\}$ and each operation is defined with a parameter triplet: $\operatorname{op} = \left< \operatorname{type} t,~\operatorname{probability} p,~\operatorname{magnitude} \lambda \right>$. %which is introduced in detail in the next section. 
Such a compositional structure allows more than one operation to be applied to the original sentence and can improve the the diversity of the synthetic instances. 
In conclusion, a policy solution consists of two kinds of knowledge to learn: the operation set and the editing parameters for each operation. 
% To search the optimal policy, we utilize the sequential Model-based Global Optimization (SMBO)~\cite{SMBO}, an efficient and widely used method in AutoML~\cite{automl}, to learn the optimal configuration of the compositional policy.
% Note that our policy is responsible for improving the quality of the augmented data, which can be reflected by the accuracy of models on the validation set.
% Therefore, we propose a novel objective function for the policy optimization.
To search the optimal policy, we propose a novel objective function and utilize the sequential Model-based Global Optimization (SMBO)~\cite{SMBO}, an efficient and widely used method in AutoML~\cite{automl}, to learn the optimal configuration of the compositional policy.

Given a target dataset, our algorithm learns an augmentation policy automatically and adaptively.
The implementation based on distributed learning frameworks can efficiently obtain promising results in several GPU hours. %, which is low-cost and fast.  
% The burden of artificial augmentation selection and parameter tuning can be substantially reduced using this algorithm. 
% In low-resource and class-imbalanced regimes of six popular datasets, TAA significantly improves the generalization ability of deep neural networks like BERT and effectively boosts text classification performance, which outperforms other baselines and prove the effectiveness of our approach. 
To summarize, our contribution is two-fold:% three-fold:
\begin{itemize}
    \item We present a learnable and compositional framework for data augmentation. Our proposed algorithm automatically searches for the optimal compositional policy, which improves the diversity and quality of augmented samples.
    \item In low-resource and class-imbalanced regimes of six benchmark datasets, TAA significantly improves the generalization ability of deep neural networks like BERT and effectively boosts text classification performance.
\end{itemize}

\section{Text AutoAugment}
\label{sec:Algo}

In this section, we introduce the Text AutoAugment framework to search for the optimal augmentation policy automatically.
Our augmentation policy is composed of various operations and forms a hierarchical structure, which is first detailed in Section~\ref{subsec:Search Space}. 
% We first detail such a compositional policy in Section~\ref{subsec:Search Space}. 
Then we give an overview of the proposed learnable augmentation method and establish a global objective function for it~(Section~\ref{subsec:Learnable Data Augmentation}). 
The specific policy optimization algorithm is presented at last~ (Section~\ref{subsec:Augmentation Policy Optimization}).

\subsection{Compositional Augmentation Policy}
\label{subsec:Search Space}

To generate a much broader set of augmentations, we introduce a compositional policy instead of a single operation. 
Figure~\ref{Fig:augmentation policy} shows the structure and usage of the policy $\mathcal{P}$. 
Specifically, our policy is a set of $N$ various editing operations: 
$$\mathcal{P}=\left\{ \mathcal{O}_1, \cdots, \mathcal{O}_i, \cdots, \mathcal{O}_N \right\}$$
The operation $\mathcal{O}$ is an atomic component and is responsible for applying an editing transformation on a text $x$ to synthesize an augmented instance $x_\text{aug}$.
Each operation $\mathcal{O}_i$ is defined as a triplet: 
$$\mathcal{O}_i = \left< t_i, p_i, \lambda_i \right>$$ 
with three parameters: 
(1) Type $t\in\{$\emph{Random Swap, Random Delete, TF-IDF Insert, TF-IDF Substitute, WordNet Substitute}$\}$.\footnote{Please refer to Appendix~\ref{appsec:editing operations} for the detailed description of each type.}
(2) Probability $p \in \left[0,1\right]$ of being applied.
(3) Magnitude $\lambda \in \left[0,0.5\right]$ which determines the proportion of the words to be edited.

The augmented text after an operation can be formulated as:
\begin{equation}
\small
    \mathcal{O}_i\left(x;\; \left<t_i,p_i,\lambda_i\right>\right)=\begin{cases}
        \mathcal{O}_i\left(x;\left<t_i,\lambda_i\right>\right) & \text{with} \; p_i   \\
        x                                 & \text{with} \; 1-p_i
    \end{cases}
    \label{equ:operation}
\end{equation}
It means that for a given text $x$, it has probability $p_i$ to be applied with the operation $\mathcal{O}_i$ for synthesizing the transformed text $x_\text{aug}=\mathcal{O}_i\left(x;\left<t_i,\lambda_i\right>\right)$ under the magnitude $\lambda_i$, while has probability $1-p_i$ to remain identical, i.e., $x_\text{aug}=x$.

To further increase the diversity of augmented data and enlarge the support of the training distribution, our policy applies more than one operation to the original sentence in a recursive way. 
Specifically, we randomly sample $N^*$ editing operations from the policy $\mathcal{P}$ and apply them to a given text consecutively. 
The $N^*$ atomic operations can be combined as a compositional operation with the recursive depth equals $N^*$. 
For example, when $N^*=2$ and the sampled operations are $[\mathcal{O}_i, \mathcal{O}_j]$, the final augmented instance can be denoted as:
$$x_\text{aug}=\mathcal{O}_j\left( \mathcal{O}_i \left( x;\left<t_i,p_i, \lambda_i  \right>\right); \left<t_j,p_j, \lambda_j \right>\right).$$

In other words, one policy can synthesize up to $A_{N}^{N^*} \times 2^{N^*}$ augmented instances, where $A$ denotes a permutation. 
Note that our policy is determined by only $N \times 3$ parameters, which does NOT cause the problem of search space explosion. 
As discussed before, the setting of these parameters in a policy has a great impact on the quality of augmented data and the model performance, which motivates us to devise a learnable framework for automatically selecting the optimal parameters instead of a naive grid search or cumbersome manual tuning.
% The choice of these parameters determines the final performance of augmentation policy. 
% In the next subsection, we will show how to optimize the parameters and learn the optimal policy.

% \subsection{Parameterized Data Augmentation}
\subsection{Learnable Data Augmentation Policy}
\label{subsec:Learnable Data Augmentation}
In this subsection, we first review the data augmentation and model training with the traditional objective function. We then propose a new objective function to learn the optimal policy for obtaining a augmented dataset with higher quality and improving the model performance.
% To learn the optimal policy for obtaining a augmented dataset with higher quality and improving the model performance, we propose a new objective function. 
% Before that, we first review the data augmentation and model training with the traditional objective function.
% Consider text classification task under usual supervised learning setting. 

Given an input space $\mathcal{X}$ and output space $\mathcal{Y}$ of the text classification task,
% a model with parameters $\theta$ needs to learn a mapping $f: \mathcal{X} \rightarrow \mathcal{Y}$ from an input text $x\in\mathcal{X}$ to a target $y\in\mathcal{Y}$.
a model $f$ is responsible for learning a mapping from input texts $x\in\mathcal{X}$ to target labels $y\in\mathcal{Y}$.
In some scenarios, the training set $\mathcal{D}_\text{train}$ is extremely small or imbalanced, which leads to a large generalization error on the test set. 
Therefore, data augmentation is incorporated as an implicit regularizer~\cite{explicit-regularization} to help models learn better patterns~\cite{Transformation-Invariance} and further improve the generalization ability. 
Let $\mathcal{D}_\text{aug}(\mathcal{P})$ be the augmented dataset containing both training set and the synthetic data generated by the policy $\mathcal{P}$, 
the loss function $\mathcal{L}$ of model training on the augmented dataset can be formulated as a sum of instance-level loss $l$ such as cross-entropy:
\begin{equation}
    \mathcal{L} = \sum_{\mathclap{(x,y)\in \mathcal{D}_\text{aug}(\mathcal{P})}} l\left(f\left(x\right), y\right).
    \label{equ:structural and data-based risk}
\end{equation}

In traditional methods, the parameters of data augmentation are preset before training then tuned on the validation set $\mathcal{D}_\text{val}$, which is similar to the paradigm of hyper-parameters tuning. 
% In order to learn augmentation policy automatically and find the best policy $\mathcal{P}$, we refer to the \emph{Combined Algorithm Selection and Hyper-parameter (CASH)} optimization problem \cite{Auto-WEKA, Auto-sklearn} in AutoML. %and propose a new algorithm named Text AutoAugment for augmentation policy optimization. 
For this reason, we cast the problem of augmentation policy optimization as the \emph{Combined Algorithm Selection and Hyper-parameter (CASH)} optimization problem \cite{Auto-WEKA, Auto-sklearn} in AutoML~\cite{automl}.
Formally, let $\mathbb{F}$ and $\mathbb{P}$ be the search space of models and policies, respectively. 
Each model $f$ is trained on $\mathcal{D}_\text{aug}(\mathcal{P})$ augmented by the policy $\mathcal{P}$. 
% We propose a novel metric to measure the loss of a policy $\mathcal{P}$ and a model $f$:
We propose a novel metric to measure the loss of a policy and a model:
% loss for optimizing P and f
% \begin{equation}
%     \mathcal{J} \left( f_\mathcal{P}, \mathcal{D}_{\text{val}} \right) = \sum_{\mathclap{(x,y)\in \mathcal{D}_{\text{val}}}} l\left(f_\mathcal{P} \left(x\right), y\right)
% \end{equation}
\begin{equation}
    \mathcal{J} \coloneqq \mathcal{J}\left( f, \mathcal{D}_\text{aug}(\mathcal{P}), \mathcal{D}_\text{val} \right)
\label{equ:total loss}
\end{equation}
Here, $\mathcal{J}\left( f, \mathcal{D}_\text{aug}(\mathcal{P}), \mathcal{D}_\text{val} \right)$ denotes the loss that the model $f$ achieves on the validation set $\mathcal{D}_\text{val}$ after trained on the augmented set $\mathcal{D}_\text{aug}(\mathcal{P})$. 
It quantifies the generalization ability of the model after augmentation thus reflects the quality of the augmented data.
Consequently, the objective function of our policy optimization can be written as
\begin{equation}
    f^*,\mathcal{P}^*=\arg\min_{ \substack{f\in \mathbb{F} \\ \mathcal{P} \in \mathbb{P}}} \mathcal{J}(f, \mathcal{D}_\text{aug}(\mathcal{P}), \mathcal{D}_\text{val}).
    \label{equ:total object}
\end{equation}
Through minimizing the validation loss of the model, % and maximizing the Distinct-N of the augmented data, 
we can find the optimal configuration of the policy $\mathcal{P}$ and improve the quality of the augmented data.

\begin{figure}[t!]
    \centering
    \includegraphics[width=0.43\textwidth]{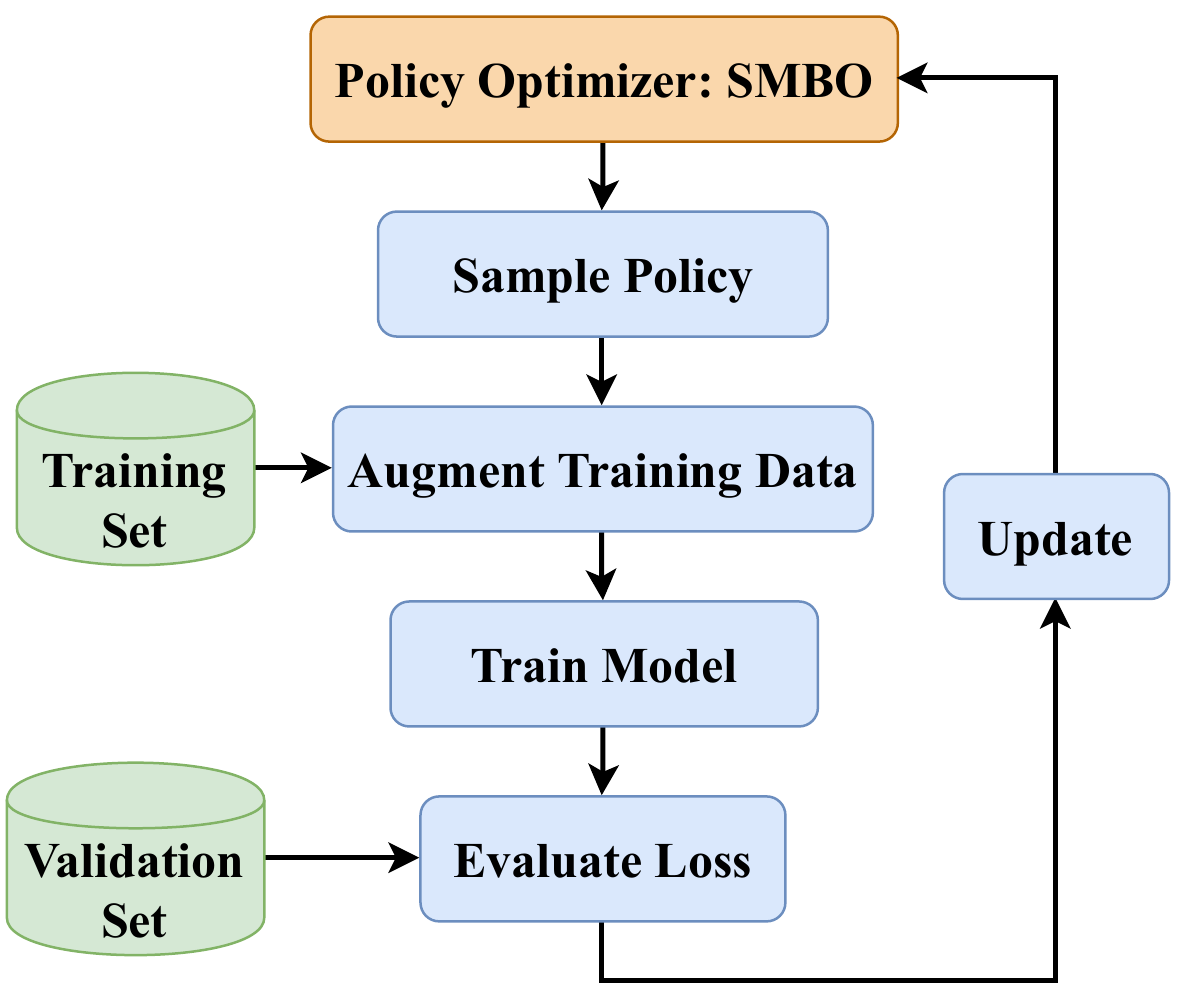}
    \caption{The overview of the optimization procedure in the Text AutoAugment algorithm. In each iteration, the optimizer samples a policy and trains a corresponding model on the augmented training set. After that, the loss on the validation set is calculated to update the policy optimizer then execute the next iteration.}
    \label{Fig:Algorithm}
\end{figure}

\subsection{Augmentation Policy Optimization}
\label{subsec:Augmentation Policy Optimization}
% In this section, we introduce the specific way to optimize our objective function in Eq.~\ref{equ:total object}. 
Like the optimization of hyper-parameters of a deep learning model, our objective function for policy optimization depends on the validation loss, % and Distinct-N, 
which can not be solved by gradient-based methods such as back-propagation. 
To tackle this problem, we introduce \emph{Sequential Model-based Global Optimization} (SMBO)~\cite{SMBO}, a kind of Bayesian optimizer widely used in AutoML, to our TAA framework.
Our optimization procedure is illustrated in Figure~\ref{Fig:Algorithm}. 
In a nutshell, the SMBO optimizer builds a probability model of the objective function as a surrogate and uses it to samples the most promising policy, then evaluates the policy in the true objective function. 
In practice, we use Tree-structured Parzen Estimator (TPE)~\cite{SMBO} as a surrogate agent $M$ to model the function between the policy and the objective loss in Eq.~\ref{equ:total loss}. 
This process is carried out in an iterative manner. 
At each iteration, the SMBO optimizer samples an augmentation policy $\mathcal{P}_i$ to synthesize the augmented set $\mathcal{D}_\text{aug}\left(\mathcal{P}_i\right)$, and then trains a model $f_i$ based on it. %by minimizing the loss in Eq.~\ref{equ:structural and data-based risk}. 
The loss $\mathcal{J}_i$ of the policy $\mathcal{P}_i$ is then evaluated by Eq.~\ref{equ:total loss} and merged into the observation history $\mathcal{H}$ to help update the surrogate model. 

The specific updating procedure of TPE is introduced in Appendix~\ref{appsec:SMBO}. 
After that, we employ the following Expected Improvement (EI) criterion as an acquisition function to sample the next promising policy:
\begin{equation}
% \begin{aligned}
%     % \resizebox{.9\hsize}{!}{$\text{EI}(\phi,M)=\mathbb{E}\left[\min\left(\mathcal{L}(\theta' \mid  \mathcal{D}_{\text{aug}}(\phi))-\mathcal{L}^{\dagger}, 0\right)\right]$}
%     \text{EI} (\phi, & M) = \\
%     \mathbb{E} & \left[ \min \left( \mathcal{L}(F, \mathcal{D}_\text{aug}(\phi), \mathcal{D}_\text{val}) - \mathcal{L}^{\dagger}, 0 \right)\right]
% \end{aligned}
\mathrm{EI}\left(\mathcal{P}\right) = \mathbb{E} \left[ \max \left( \mathcal{J}^{\dagger} - \mathcal{J}, 0 \right)\right]
\label{equ:EI}
\end{equation}
Here, $\mathcal{J}^{\dagger}$ is a threshold calculated by the observation history and the surrogate model.
Eq.~\ref{equ:EI} is the expectation under the surrogate model that the loss $\mathcal{J}$ of a policy will exceed (negatively) the threshold $\mathcal{J}^{\dagger}$. 
Note that our target is to find the policy that minimizes the loss $\mathcal{J}$, the policy that maximizes the expected improvement will be chosen in the next iteration. 
% During the phase of policy optimization, we explore the augmentation policy iteratively. 
% At each iteration, SMBO samples the policy $\mathcal{P}_i$ that maximizes EI. 
% Then we train a model $f_i$ on augmented set $\mathcal{D}_\text{aug}(\mathcal{P}_i)$ and evaluate it on actual validation set.
% After that, we update the surrogate model and start the next exploration.
The TAA framework is summarized in Algorithm~\ref{alg:TAA algorithm}.\footnote{Due to limited space, please refer to Appendix~\ref{appsec:SMBO} for the detailed description of the surrogate model update and the EI calculation.}
% In practice, we use \texttt{HyperOpt}~\cite{HyperOpt} library from \texttt{Ray}~\cite{Ray} for the parallelized implementation.

% \floatname{algorithm}{Algorithm}
% \renewcommand{\algorithmicrequire}{\textbf{Input:}}
% \renewcommand{\algorithmicensure}{\textbf{Output:}}
% \begin{algorithm}
%     \centering
%     \caption{TAA Algorithm}
%     \begin{algorithmic}[1]
%         \Require Search space of $\phi$: $S$
%         \Require Search space of $F$: $\mathcal{F}$
%         \Require Number of iterations $N$
%         \Require Training set $D_{\text{train}}$, validation set $D_{\text{val}}$

%         \State{$\mathcal{H}\leftarrow \emptyset$}\Comment{Initialize observation history}
%         \For {$i=1$ to $N$}
%         \State{$\phi_i \leftarrow \arg\max_\phi \text{EI}\left(\phi, M_{i-1}\right)$}\Comment{See Eq.~\ref{equ:EI}}
%         \State{Train $F_i$ through Eq.~\ref{equ:structural and data-based risk}}
%         \State{Evaluate $\mathcal{L}(F_i,D_\text{aug}(\phi_i), D_\text{val})$ in Eq.~\ref{equ:total object}}
%         \State{$\mathcal{H} \leftarrow \mathcal{H} \cup \left(\phi_i, \mathcal{L}(F_i, D_\text{aug}(\phi_i), D_\text{val})\right)$}
%         \State{Update surrogate model $M_i$ to fit $\mathcal{H}$}
%         \EndFor
%         \Ensure Learned model $F^*$
%         \Ensure Best augmentation policy $\phi^*$
%     \end{algorithmic}
%     \label{alg:TAA algorithm}
% \end{algorithm}

\newcommand\mycommfont[1]{\small\ttfamily{#1}}
\SetCommentSty{mycommfont}
\SetKwInput{KwInput}{Input}
\begin{algorithm}[t!]
\small
\DontPrintSemicolon
\SetAlgoLined
% \KwInput{Search space of $\phi$: $S$}
% \KwInput{Search space of $F$: $\mathcal{F}$}
\KwInput{Number of iterations $T$}
\KwInput{Training set $D_{\text{train}}$}
\KwInput{Validation set $D_{\text{val}}$}
$\mathcal{H}\leftarrow \emptyset$ \tcp*{Initialize history}
\For{$i=1$ to $T$}{
    $\mathcal{P}_i \leftarrow \arg\max_\mathcal{P} \mathrm{EI}\left(\mathcal{P}\right)$ \tcp*{Eq.~\ref{equ:EI}}
    Train $f_i$ \tcp*{Eq.~\ref{equ:structural and data-based risk}}
    Evaluate $\mathcal{J}_i$ \tcp*{Eq.~\ref{equ:total loss}}
    $\mathcal{H} \leftarrow \mathcal{H} \cup \left(f_i, \mathcal{P}_i, \mathcal{J}_i \right)$ \;
    Update surrogate model $M$ to fit $\mathcal{H}$ \;
}
$f^*,\mathcal{P}^*=\arg\min_{ \substack{f\in \mathcal{H} \\ \mathcal{P} \in \mathcal{H}}} \mathcal{J}$\tcp*{Eq.~\ref{equ:total object}}
\Return{$f^*$ and $\mathcal{P}^*$}
% \Return{The optimal augmentation policy $\mathcal{P}^*$}
\caption{TAA Algorithm}
\label{alg:TAA algorithm}
\end{algorithm}

\section{Experiments}
\label{sec:Exp}
% For empirical evaluation, TODO
\subsection{Benchmark Datasets}
\label{subsec:Benchmark Datasets and Base Model}
We conduct experiments on six benchmark datasets, including IMDB~\cite{IMDB}, SST-2, SST-5~\cite{SST}, TREC~\cite{TREC}, YELP-2 and YELP-5~\cite{YELP}. The statistics of the datasets are listed in Appendix~\ref{appsec:datasets statistics}.

\subsection{Baselines}
\label{subsec:Comparison Augment Methods}
We compare our TAA method with the following representative baselines:

\medskip
\noindent\textbf{Back Translation (BT)}~\cite{improving-MT, enhancement-MT}.
% is a sentence-level operation.
We utilize WMT’19 English-German translation models~\cite{WMT19-En-Ge} based on  Transformer~\cite{transformer} %implemented by fairseq~\cite{ott2019fairseq} 
to translate the text from English to German,\footnote{\url{https://github.com/pytorch/fairseq/blob/master/examples/translation/README.md}} and then translate it back to English. % with the corresponding German-English model.
%   The model checkpoints are the same as that in WMT’19\footnote{\url{https://github.com/pytorch/fairseq/blob/master/examples/translation/README.md}}.
We use random sampling for decoding as recommended by~\cite{uda, understanding-bt}, and set the temperature to 0.8 to generate more diverse paraphrases.

\medskip
\noindent\textbf{Contextual Word Substitute (CWS)}~\cite{contextual-augmentation}. 
% is an improved version of traditional synonym replacement method. It 
CWS masks some words in the original text stochastically, then predicts new words for substitution using a label-conditional language model (LM). 
% Following the original paper, we utilize a bi-directional LSTM-RNN to implement the label-conditional LM. 
The proportion $\alpha$ of words to be masked and predicted is 0.15.

\medskip
\noindent\textbf{Easy Data Augmentation (EDA)}~\cite{eda}. Given a text from the training set, EDA randomly applies one of four simple editing operations including Synonym Replacement, Random Insertion, Random Swap, and Random Deletion.
%   Given a text from training set, EDA randomly choose and apply one of the above four operations.
%   The unique parameter of EDA is $\alpha$, which indicates the proportion of words to be modified in the text. 
We set the proportion of words to be edited to 0.05 according to the recommendations in the paper.

\medskip
\noindent\textbf{Learning Data Manipulation (LDM)} \cite{learning-data-manipulation}.
LDM is also a learnable data augmentation method, but uses only one operation: aforementioned CWS.
LDM establishes a reinforcement learning system to  iteratively optimize the parameters of the task model and label-conditional LM. 
%   Note that $\alpha$ in LDM is also set to 0.15 arbitrarily. 
% \item {\bf Random}. For comprehensive comparison, we adopt the same search space described in Section~\ref{subsec:Search Space}, but do not conduct any policy optimization.
%       For every text in the training set, a random policy is sampled to synthesize the augmented data.

The main characteristics of the baselines are summarized in Table~\ref{tb:baselines}. 

\subsection{Experimental Settings}
\label{subsec:Training Setting}
% During the phase of policy optimization and final training, we both use the BERT model with the same training setting.
We choose large-scale pre-trained BERT (base, uncased version)~\cite{bert} as the backbone model. 
% For the BERT model, w
We adopt the Adam optimizer~\cite{Adam} and a linear warmup scheduler with an initial learning rate of 4e-5. 
The training epoch is 20 for TREC and 10 for the reset datasets. We pick the best checkpoint according to the validation loss. 
% In order to avoid underfitting or overfitting caused by the different sizes of the original and augmented training set, we apply early stopping with the patience of 2 according to the valid loss.
Note that the validation loss is also the criterion for policy optimization, so we split out two validation sets with the same size during the phase of policy optimization. 
One is used for evaluating the model checkpoints, and the other is for evaluating the sampled policy.
All experiments are conducted with 8 Tesla P40 GPUs. % and 100GB RAM.
% Referring to the setting in
Following previous works~\cite{learning-data-manipulation, uda, eda}, we verify the performance of all data augmentation baselines in two special data scenarios:

\medskip
\noindent\textbf{Low-resource Regime}.
For every dataset in Section~\ref{subsec:Benchmark Datasets and Base Model}, %Table~\ref{tb:dataset detail}, 
we constrain the amount of available labeled data by sub-sampling a smaller training and validation set.
% Let $N_c$ be the number of classes in a dataset.
In order to maintain the distribution of the original dataset, we apply \emph{Stratified ShuffleSplit}~\cite{stratified-shuffling} to split the training and validation set.
%with size of $N_c*40$ and $N_c*30$ respectively.
% Therefore, IMDB, SST-5 and TREC has 80, 200, 240 labeled training samples separately, which poses significant challenges for learning.
The final datasets IMDB, SST-5, TREC, YELP-2, and YELP-5 have 80, 200, 120, 80, 200 labeled training samples, respectively, which pose significant challenges for learning a well-performing classifier. 
The number of validation samples is 60, 150, 60, 60, 150, respectively. 
In low-resource regime, we introduce a parameter $n_\text{aug}$, representing the magnification of augmentation. For example, $n_\text{aug}=16$ means that we synthesize 16 samples for each given sample. 
%In other words, the size of the augmented training set is 16 times that of the original one. 

\begin{table}[t!]
\centering
\small 
% \scalebox{0.9}{
\setlength{\tabcolsep}{1mm}{
\begin{tabular}{lccc}
\toprule
Baselines          & Learnable & Compositional & Op. Level    \\ \midrule
BT                 & \xmark        & \xmark           & Sentence \\
CWS                & \xmark        & \xmark            & Word     \\
EDA                & \xmark        & \cmark          & Word     \\
LDM                & \cmark       & \xmark             & Word     \\
\textbf{TAA (Ours)} & \cmark       & \cmark           & Word     \\ \bottomrule
\end{tabular}}
% }
\caption{Comparison of baselines.}
\label{tb:baselines}
\end{table}

\begin{table*}[t!]
\centering
\begin{adjustbox}{max width=0.95\textwidth}
% results from tencent
    % \begin{tabular}{lcccccc}
    %     \toprule
    %     Method   & IMDB (80)       & SST-5 (200)     & TREC (120)      & YELP-2 (80)    & YELP-5 (200)  & Average    
    %     \\ \midrule
    %     % Training  & 95.08\small$\pm$4.51 & 65.37\small$\pm$9.76 & 98.06\small$\pm$1.34 & 98.33\small$\pm$1.69 & 74.97\small$\pm$12.44 & \small$\pm$ \\ \midrule
    %     No Aug  & 64.47\small$\pm$5.14 & 34.67\small$\pm$4.68 & 69.31\small$\pm$5.37 & 73.59\small$\pm$3.13 & 37.42\small$\pm$3.65 & 55.89\small$\pm$4.39 \\
    %     CWS  & 67.67\small$\pm$2.57 & 34.57\small$\pm$2.70 & 74.31\small$\pm$5.01 &  75.89\small$\pm$3.32  & 39.96\small$\pm$3.55 & 58.48\small$\pm$3.43         \\ 
    %     EDA  & 71.75\small$\pm$5.80 & 37.66\small$\pm$2.29 & 70.38\small$\pm$5.25 & 75.70\small$\pm$7.64   & 41.16\small$\pm$3.87 & 59.33\small$\pm$4.97     \\
    %     LDM  & 70.39\small$\pm$3.61 & 34.84\small$\pm$2.77 & 75.33\small$\pm$6.77 &  78.79\small$\pm$3.01  &  40.82\small$\pm$3.99 & 60.03\small$\pm$4.03 \\ 
    %     BT   & 70.67\small$\pm$4.04 & 38.20\small$\pm$4.60 & 78.08\small$\pm$4.07 &  78.97\small$\pm$2.95  &  42.19\small$\pm$2.53  & 61.62\small$\pm$3.64      \\ 
    %     \textbf{TAA(Ours)}  & \textbf{75.12}\small$\pm$5.20 & \textbf{39.43}\small$\pm$3.18 & \textbf{79.99}\small$\pm$4.21 &  \textbf{80.67}\small$\pm$2.89   & \textbf{44.87}\small$\pm$2.67 & \textbf{64.02}\small$\pm$3.63 \\ \bottomrule
    % \end{tabular}
% results from ali & lanco
\begin{tabular}{lcccccc}
        \toprule
        Method   & IMDB (80)       & SST-5 (200)     & TREC (120)      & YELP-2 (80)    & YELP-5 (200)  & Average    
        \\ \midrule
        % Training  & 95.08\small$\pm$4.51 & 65.37\small$\pm$9.76 & 98.06\small$\pm$1.34 & 98.33\small$\pm$1.69 & 74.97\small$\pm$12.44 & \small$\pm$ \\ \midrule
        No Aug  & 64.74\small$\pm$3.41 & 36.14\small$\pm$3.99 & 69.31\small$\pm$5.37 & 73.87\small$\pm$4.22 & 36.62\small$\pm$4.67 & 56.14\small$\pm$4.33 \\
        CWS  & 73.44\small$\pm$3.56 & 39.22\small$\pm$2.66 & 74.31\small$\pm$5.01 &  78.87\small$\pm$3.96  & 43.05\small$\pm$2.18 & 61.77\small$\pm$3.47         \\ 
        EDA  & 73.93\small$\pm$1.88 & 39.72\small$\pm$1.93 & 71.59\small$\pm$4.39 & 77.46\small$\pm$4.78   & 43.17\small$\pm$2.76 & 61.17\small$\pm$3.15     \\
        LDM  & 70.39\small$\pm$3.61 & 40.25\small$\pm$1.48 & 75.33\small$\pm$4.77 &  79.70\small$\pm$2.85  &  43.85\small$\pm$0.96 & 61.90\small$\pm$2.73 \\ 
        BT   & 74.13\small$\pm$3.10 & 39.40\small$\pm$3.87 & 78.08\small$\pm$4.07 &  78.97\small$\pm$2.95  & 42.19\small$\pm$2.53  & 62.55\small$\pm$3.30      \\ 
        \textbf{TAA(Ours)}  & \textbf{75.68}$^*$\small$\pm$3.27 & \textbf{40.28}\small$\pm$1.80 & \textbf{81.47}$^{**}$\small$\pm$3.87 &  \textbf{81.75}$^{*}$\small$\pm$3.57   & \textbf{45.29}$^{*}$\small$\pm$1.76 & \textbf{64.89}\small$\pm$2.85 \\ \bottomrule
    \end{tabular}
\end{adjustbox}
    \caption{Test accuracy (\%) with standard deviation of different augmentation methods in low-resource regime. IMDB (80) means the number of training samples after sub-sampling is 80. $^*$ and $^{**}$ indicate statistically significant (p < .05 and p < .01) improvements over the best baseline. Here $n_\text{aug}=16$.}
    \label{tb:low data accuracy}
\end{table*}

\medskip
\noindent\textbf{Class-imbalanced Regime}.
% We simulate the class-imbalanced regime on binary sentiment classification datasets.
% , it can be easily extended to other types of tasks. 
For binary sentiment classification datasets IMDB and SST-2, 
we sub-sample the training samples from an imbalanced class distribution. 
% we also sub-sample a training set and validation set but in a different way. 
After sub-sampling, the negative class of the training set has 1000 samples, while the positive class has only 20/50/100 training samples respectively in three experiments. 
% That is, the data amount of category 2 is 50/20/10 times that of another category 1. 
In this regime, we only augment the samples in positive class by 50/20/10 times, so that the final data amount of the positive and negative class are the same. We set up another baseline {\bf Over-Sampling (OS)} for this scenario, which over-samples the training samples in the positive class by 50/20/10 times. Following the settings of the previous work~\cite{learning-to-reweight, learning-data-manipulation}, the validation set for training and policy optimization is balanced.

In both low-resource and class-imbalanced regimes, the test set for final evaluation is \textbf{balanced} and \textbf{intact} without any reduction, which indicates that accuracy is a fair metric for evaluation. 
The size of policy $N$ and the number of the operations sampled for each text $N^*$ is 8 and 2, respectively. 
For a fair comparison, we use 3 random seeds for data sub-sampling and run 5 times under each seed.
Finally, we report the results on the average of 15 runs with standard deviation (denoted as $\pm$).

\subsection{Main Results}
\label{subsec:Evaluation Results}
We train the model on the training set augmented by baselines in Section~\ref{subsec:Comparison Augment Methods}, then conduct evaluations on the balanced and intact test set.

% \subsubsection{Low Data Regime}
% \label{subsec:Low Data Regime}
\medskip
\noindent\textbf{Low-resource Regime}. 
Table~\ref{tb:low data accuracy} shows the test accuracy of TAA on five datasets. 
In the low-resource regime, the model suffers from a severe overfitting problem. For example, the accuracy on the sub-training set of TREC is as high as 98.06\%$\pm$1.34\%, while only obtains 69.3\%$\pm$5.4\% on the test set.
Several kinds of data augmentation baselines have greatly improved the generalization ability of the model. 
% Interestingly, EDA algorithm without optimization even performs better than the learnable algorithm LDM on the IMDB and SST-5, which shows the power of the diverse augment operations. 
As shown in Figure~\ref{Fig:imdb_m}, however, the performance of the augment operations is very sensitive to their parameters like the magnitude. The heuristic-based approaches such as EDA and CWS are likely to be trapped in sub-optimum because of the manually parameters setting. 
On the contrary, our Text AutoAugment algorithm with a learnable and compositional policy outperforms all the baselines by a considerable margin. Compared to the model without augmentation, TAA boosts the accuracy by \textbf{8.8\%} averagely. 
Note that it does not cost too much time of computation to achieve such performance. When the number of iterations $T$ (see Algorithm~\ref{alg:TAA algorithm}) is 200, it only requires 4.24$\pm$0.10 GPU hours on 8 NVIDIA Tesla P40 GPUs to finish the optimization. 

\medskip
\noindent\textbf{Class-imbalanced Regime}. 
To verify that TAA can also improve the performance of the model in class-imbalanced regime, we conduct experiments based on the settings in Section~\ref{subsec:Training Setting}. 
As illustrated in Table~\ref{tb:imbalanced data accuracy}, the over-sampling method alleviates the overfitting problem to some extent but is not as efficient as augmentation baselines. In contrast, our TAA boosts the test accuracy by an average of about \textbf{9.7\%}, which surpasses other algorithms. %{\color{red} TODO}

In both low-resource and class-imbalanced regimes, our TAA framework based on \emph{multiple learnable} operations surpasses BT \& CWS based on \emph{single unlearnable} operation, EDA based on \emph{multiple unlearnable} operations, and LDM based on \emph{single learnable} operation.
% The policies searched in the low-resource regime can also be directly applied to the \textbf{full} dataset to improve the model performance by 0.4\% on average. Please refer to Appendix E for the detailed results.

\begin{figure}[t!]
    \centering
    \includegraphics[width=0.9\linewidth]{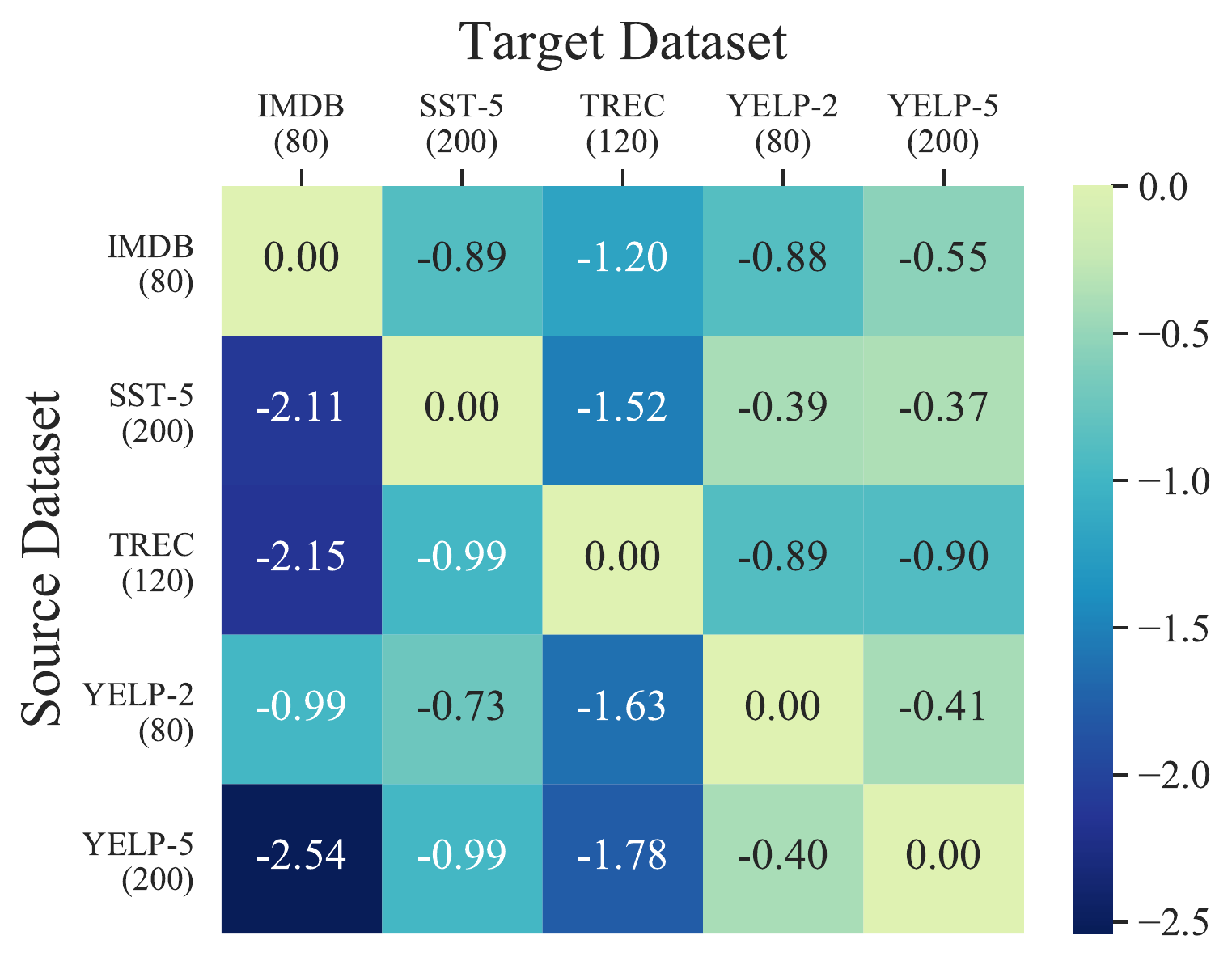}
    \caption{Transferability of the policies searched by TAA from one source dataset to other target datasets. The number denotes the performance degradation when the policy is transferred to the target dataset.}
    \label{Fig:transfer_taa}
\end{figure}

\section{Analysis}
\label{sec:Analysis}
We analyse the learned policies on four aspects including transferability, scalability, magnification and structure. 
Besides, we evaluate the diversity and semantics preservation of the augmented data, and further conduct a case study. The final policies searched by our TAA algorithm are listed in Appendix~\ref{appsec:searched policy}.

\begin{table*}[t!]
\centering
% \small 
\begin{adjustbox}{max width=0.95\textwidth}
    \begin{tabular}{lccccccc}
        \toprule
        Method          & \begin{tabular}[c]{@{}c@{}}SST-2\\ (20:1000)\end{tabular} & \begin{tabular}[c]{@{}c@{}}SST-2\\ (50:1000)\end{tabular} & \begin{tabular}[c]{@{}c@{}}SST-2\\ (100:1000)\end{tabular} & \begin{tabular}[c]{@{}c@{}}IMDB\\ (20:1000)\end{tabular} & \begin{tabular}[c]{@{}c@{}}IMDB\\ (50:1000)\end{tabular} & \begin{tabular}[c]{@{}c@{}}IMDB\\ (100:1000)\end{tabular} & Average \\ \midrule
        No Aug                & 50.17\small$\pm$1.30            & 55.01\small$\pm$6.33            & 68.01\small$\pm$8.80            & 50.05\small$\pm$0.00             & 51.84\small$\pm$4.56             & 65.81\small$\pm$7.65    & 56.82\small$\pm$4.77         \\
        OS               & 52.01\small$\pm$1.80            & 58.58\small$\pm$5.56            & 68.97\small$\pm$3.48            & 52.29\small$\pm$2.69             & 59.08\small$\pm$6.33             & 68.63\small$\pm$4.67     & 59.93\small$\pm$4.09        \\
        BT                 & 54.05\small$\pm$4.20            & 59.47\small$\pm$5.67            & 71.70\small$\pm$5.29            & 52.24\small$\pm$2.90             & 57.43\small$\pm$6.85             & 67.20\small$\pm$6.89   & 60.35\small$\pm$5.46          \\
        EDA                & 52.91\small$\pm$3.87            & 59.53\small$\pm$5.86            & 70.05\small$\pm$5.34            & \textbf{57.29}\small$\pm$5.74    & 64.09\small$\pm$7.18             & 71.28\small$\pm$4.94     & 62.53\small$\pm$5.49        \\
        CWS                & 53.09\small$\pm$1.98            & 64.62\small$\pm$5.05            & 74.67\small$\pm$3.88            & 55.58\small$\pm$3.34             & 64.37\small$\pm$3.55             & 74.33\small$\pm$4.92      & 64.44\small$\pm$3.79       \\
        \textbf{TAA(Ours)} & \textbf{56.45}$^*$\small$\pm$3.67   & \textbf{66.05}\small$\pm$4.85   & \textbf{75.12}\small$\pm$5.20   & 56.92\small$\pm$2.80             & \textbf{66.73}$^*$\small$\pm$4.20    & \textbf{77.87}$^*$\small$\pm$3.18  & \textbf{66.52}\small$\pm$3.98  \\ \bottomrule
    \end{tabular}
    \end{adjustbox}
    % \caption{Test accuracy (\%) with standard deviation of different augmentation methods in class-imbalanced regime. SST-2 (20:1000) means the positive class of SST-2 has 20 training samples while the negative class has 1000 training samples. Similar for other notations.}
    \caption{Test accuracy (\%) with standard deviation in class-imbalanced regime. SST-2 (20:1000) means the positive class of SST-2 has 20 training samples while the negative class has 1000 training samples. $^*$ indicates statistically significant (p < .05) improvements over the best baseline.}
    \label{tb:imbalanced data accuracy}
\end{table*}

\subsection{Transferability}
We conduct experiments to study whether the policies searched on one source dataset can be applied to other target datasets. 
As illustrated in Figure~\ref{Fig:transfer_taa}, the transfer of the policy to other datasets only results in a slight performance degeneration of 1.1\% on average and can boosts the test accuracy by 7.7\% comparing to no augmentation. 
The policies maintain better transferability when the gap of class number and sequence length\footnote{Please refer to Appendix~\ref{appsec:datasets statistics}} is smaller between the source and target datasets.
% Among them, IMDB is more sensitive to the transferred policies. We speculate that it is because the class number and sample length of imdb are quite different from other source data sets.
% the policies learned by TAA transfer well across different datasets and tasks. For example, the policy learned on IMDB leads to the absolute improvements (test accuracy) of 3.24, 9.95, 5.38, and 6.94 on SST5, TREC, YELP-2, and YELP-5, respectively.
% \input{tables/transferability}

\begin{table}[t!]
\centering
\begin{adjustbox}{max width=0.95\linewidth}
\begin{tabular}{lccc}
\toprule
Dataset & No Aug & TAA   & Improvement ($\uparrow$) \\ \midrule
IMDB    & 88.77  & \textbf{89.37} & 0.60     \\
SST-5   & 52.29  & \textbf{52.55} & 0.26     \\
TREC    & 96.40  & \textbf{97.07} & 0.67     \\
YELP-2  & 95.85  & \textbf{96.04} & 0.19     \\
YELP-5  & 65.55  & \textbf{65.73} & 0.18     \\
Average & 79.77  & \textbf{80.15} & 0.38     \\ \bottomrule
\end{tabular}
\end{adjustbox}
    \caption{Test accuracy (\%) of TAA on full datasets. Here $n_\text{aug}=4$ for IMDB, SST-5 and TREC. $n_\text{aug}=2$ for YELP-2 and YELP-5.}
    \label{tb:full-data-accuracy}
\end{table}

% \begin{table*}[t!]
% \centering
% \begin{adjustbox}{max width=0.9\textwidth}
%     \begin{tabular}{lcccccc}
%         \toprule
%         Method   & IMDB (80)       & SST-5 (200)     & TREC (120)      & YELP-2 (80)    & YELP-5 (200)  & Average    
%         \\ \midrule
%         % Training  & 95.08\small$\pm$4.51 & 65.37\small$\pm$9.76 & 98.06\small$\pm$1.34 & 98.33\small$\pm$1.69 & 74.97\small$\pm$12.44 & \small$\pm$ \\ \midrule
%         No Aug  & - & - & - & - & - & - \\
%         CWS  & - & - & - &  -  & - & -         \\ 
%         EDA  & - & - & - & -   & - & -     \\
%         LDM  & - & - & - &  -  &  - & - \\ 
%         BT   & - & - & - &  -  & -  & -      \\ 
%         \textbf{TAA(Ours)}  & - & - & - &  -   & - & - \\ \bottomrule
%     \end{tabular}
% \end{adjustbox}
%     \caption{Test accuracy (\%) with standard deviation of different augmentation methods on full dataset. Here $n_\text{aug}=16$.}
%     \label{tb:full data accuracy}
% \end{table*}

\subsection{Scalability}
In order to investigate whether the TAA policy is still helpful as the amount of training data increases, we apply the policies searched in the low-resource regime to the corresponding full datasets. 
Note that the only difference between the experimental setting here and that in the low-resource and class-imbalanced regimes is the amount of the training data. 
The test sets are intact and balanced in all three situations. 
As shown in Table~\ref{tb:full-data-accuracy}, 
% the policies can further improve model performance, but the gain may diminish as the overfitting issue becomes less serious with abundant data.
% although the overfitting issue on the full dataset becomes far less serious, 
% the learned policies can further improve the model performance by an average of 0.38.
the policies searched on a subset with only 2k training samples on average have great scalability. Such policies can improve the quality and diversity of the full training set, and further, improve the model performance by an average of 0.38.

\subsection{Impact of Augment Magnification}
\label{subsec:Impact of Augment Magnification}
\begin{figure}[t!]
    \centering
    \includegraphics[width=\linewidth]{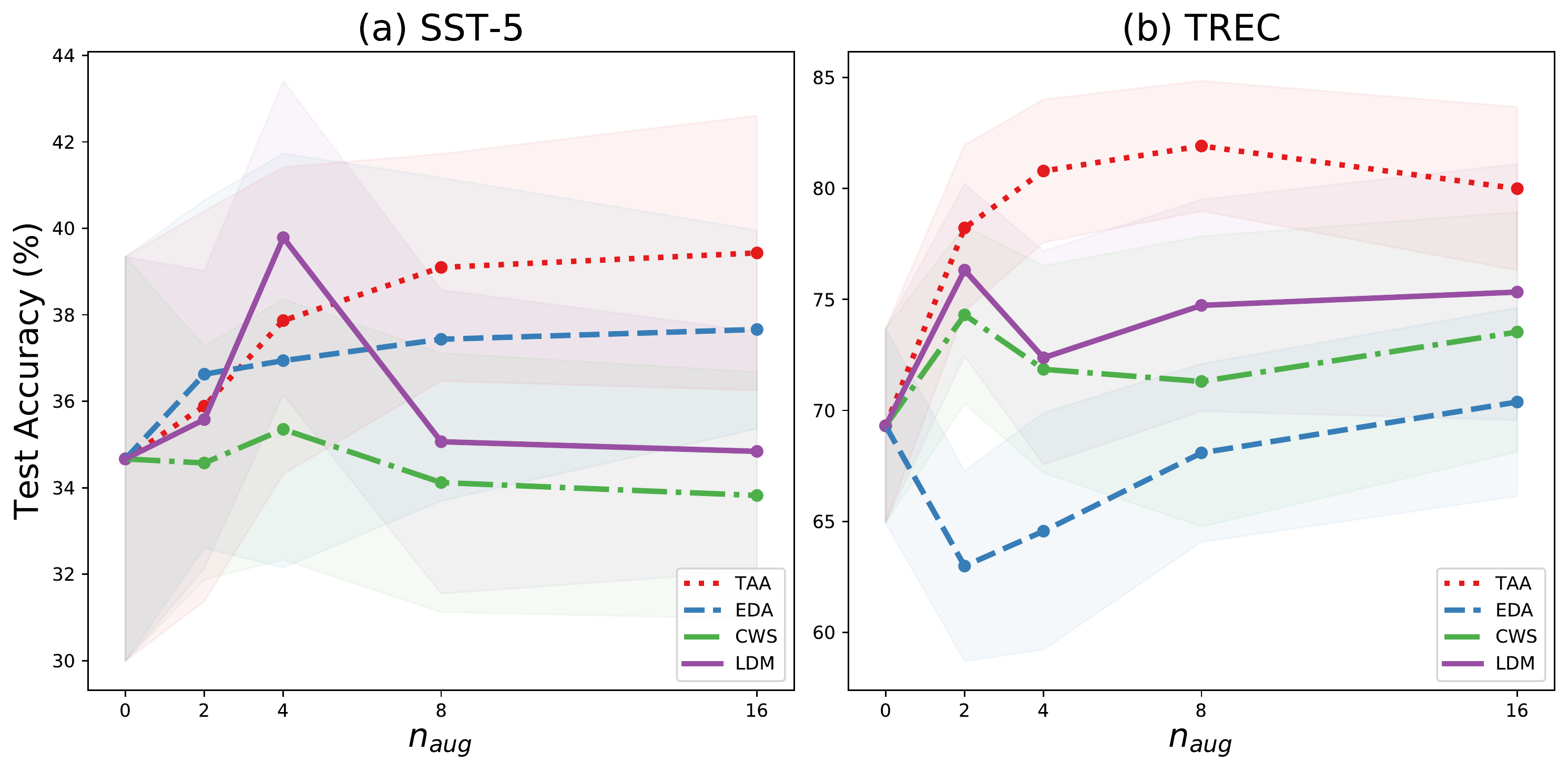}
    \caption{Test accuracy (\%) with different $n_\text{aug}$ on SST-5 and TREC, respectively. Each curve in the sub-figure denotes a kind of baseline. Shaded regions indicate standard deviation over 15 trials.}
    \label{Fig:Accuracy with different n_aug}
\end{figure}

% We analyze the impact of the augment magnification. 
Figure~\ref{Fig:Accuracy with different n_aug} shows the results of each method with different augment magnification $n_\text{aug}$. 
We find that CWS and LDM achieve their best performance on SST-5 when $n_\text{aug}=4$, but the accuracy drops sharply as $n_\text{aug}$ continues to increase. 
The performance improvement from methods with a single operation cannot scale with more augmented samples, presumably because of the lack of novel information. 
% As more augmented samples are generated, these methods based on the single operation may not be able to introduce more valid information but causes redundancy, which leads to poor performance. 
Worsely, EDA, which is based on unlearnable multiple operations, even harms the model training on TREC, due to its inappropriate parameter setting with human experience~\cite{eda}. 
In virtue of the compositional and learnable policy, the augmented data synthesized by TAA are effective and perform well in most cases.

% \begin{figure}[t!]
%     \subfigure{
%         \begin{minipage}[b]{0.48\linewidth}
%             \includegraphics[height=1.1in]{n_aug_sst5.pdf}
%             \centering
%             \\ \scriptsize{(a) SST-5}
%         \end{minipage}}
%     \subfigure{
%         \begin{minipage}[b]{0.48\linewidth}
%             \includegraphics[height=1.1in]{n_aug_trec.pdf}
%             \centering
%             \\ \scriptsize{(b) TREC}
%         \end{minipage}}
%     \caption{Test accuracy (\%) with different $n_\text{aug}$ on SST-5 and TREC, respectively. Each curve in the sub-figure denotes a kind of baseline. Shaded regions indicate standard deviation over 15 trials.}
%     \label{Fig:Accuracy with different n_aug}
% \end{figure}

\subsection{Impact of the Policy Structure}
\label{subsec:Impact of Policy Structure}
\begin{figure}[t!]
    \centering
    \includegraphics[width=0.42\textwidth]{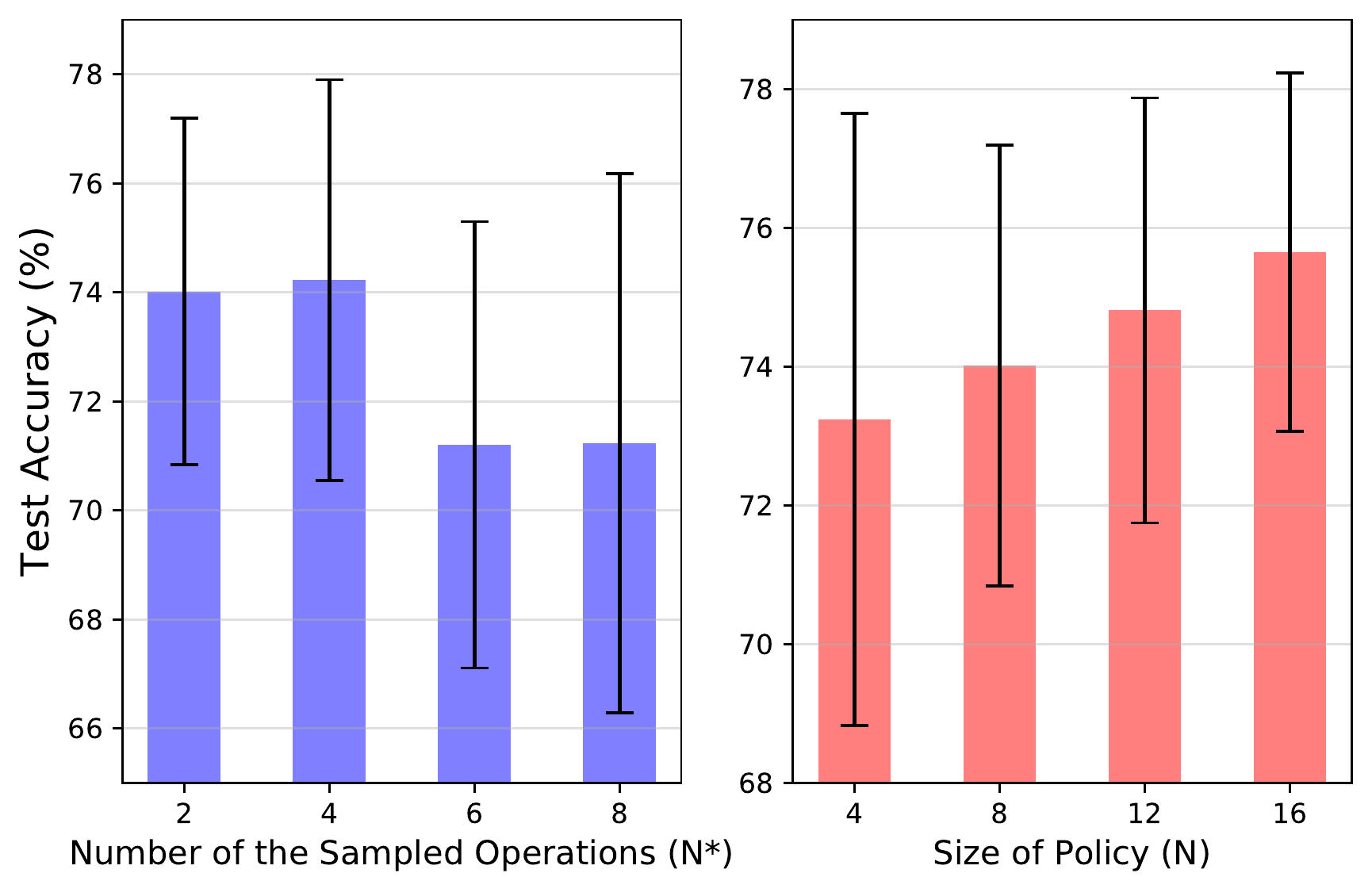}
    \caption{Performance of TAA with respect to the structure of augmentation policy. }
    \label{Fig:imdb_structure}
\end{figure}

% In order to explore the performance of TAA with respect to the structure of augmentation policy, 
% We change the size of the sub-policy and policy 
We change the size of the policy $N$ and the number of the operations sampled for each text $N^*$
and re-execute the experiments to explore the impact of the policy structure. 
The left panel in Figure~\ref{Fig:imdb_structure} shows the results with different $N^*$. %$|\mathcal{SP}|$. 
% As $|\mathcal{SP}|$ increases, 
As $N^*$ increases, the original text is likely to be applied with more operations sequentially, which causes slight damage to the performance of TAA.
While the right panel shows that the more %sub-policy 
operations a policy contains, the better TAA performs. 
Therefore, for the compositional data augmentation algorithm, it is helpful to generate more diverse samples. 
Other ablation studies on operation type and searching algorithm can be found in Appendix~\ref{appsec:ablation study}. 

\begin{table}[t!]
\centering
\small 
\begin{adjustbox}{max width=0.49\textwidth}
    \begin{tabular}{l|ccccc|c}
        \toprule
        Method   & \begin{tabular}[c]{@{}c@{}}IMDB\\ (80)\end{tabular}        & \begin{tabular}[c]{@{}c@{}}SST-5\\ (200)\end{tabular}     & \begin{tabular}[c]{@{}c@{}}TREC \\ (120)\end{tabular}      & \begin{tabular}[c]{@{}c@{}}YELP-2\\ (80)\end{tabular}    & \begin{tabular}[c]{@{}c@{}}YELP-5\\ (200)\end{tabular}  & Average    \\ \midrule
        CWS  & 0.302 & 0.731 & 0.827 &  0.390  & 0.393 &   0.529       \\ 
        EDA  & 0.304 & 0.747 & 0.839 &  0.397  & 0.399 &  0.537  \\
        % LDM  &  &  &  &    &   &  \\ 
        BT   & \textbf{0.345} & 0.745 & \textbf{0.849} &  0.439  & 0.440  &  0.564     \\ 
        \textbf{TAA(Ours)}  & \textbf{0.345} & \textbf{0.751} & 0.841 &  \textbf{0.445}   & \textbf{0.446} & \textbf{0.566} \\ \bottomrule
    \end{tabular}
\end{adjustbox}
    \caption{Distinct-2 of different augmentation methods in low-resource regimes.}
    \label{tb:distinct-n}
\end{table}

% \begin{table*}[t!]
% \centering
% \small 
% \begin{adjustbox}{max width=\textwidth}
%     \begin{tabular}{lcccccc}
%         \toprule
%         Method   & IMDB (80)       & SST-5 (200)     & TREC (120)      & YELP-2 (80)    & YELP-5 (200)  & Average    \\ \midrule
%         CWS  & 0.302 & 0.731 & 0.827 &  0.390  & 0.393 &   0.529       \\ 
%         EDA  & 0.304 & 0.747 & 0.839 &  0.397  & 0.399 &  0.537  \\
%         % LDM  &  &  &  &    &   &  \\ 
%         BT   & \textbf{0.345} & 0.745 & \textbf{0.849} &  0.439  & 0.440  &  0.564     \\ 
%         \textbf{TAA(Ours)}  & \textbf{0.345} & \textbf{0.751} & 0.841 &  \textbf{0.445}   & \textbf{0.446} & \textbf{0.566} \\ \bottomrule
%     \end{tabular}
% \end{adjustbox}
%     \caption{Distinct-2 of different augmentation methods in low-resource regime. IMDB (80) means the number of training samples after sub-sampling is 80. Here $n_\text{aug}=16$ and number of iterations $T$ is 200.}
%     \label{tb:distinct-n}
% \end{table*}
\begin{table*}[t!]
\centering
\small 
\begin{adjustbox}{max width=\textwidth}
    \begin{tabular}{lcccccc}
        \toprule
        Method   & IMDB (80)       & SST-5 (200)     & TREC (120)      & YELP-2 (80)    & YELP-5 (200)  & Average    \\ \midrule
        CWS  & 0.698 & 0.796 & 0.784 &  0.739  & 0.736 &    0.751      \\ 
        EDA  & 0.738 & 0.665 & 0.610 &  0.785  & \textbf{0.785} &  0.717  \\
        % LDM  &  &  &  &    &   &  \\ 
        BT   & 0.738 & \textbf{0.799} & \textbf{0.892} & 0.741   & 0.747  & \textbf{0.783}      \\ 
        \textbf{TAA(Ours)}  & \textbf{0.771} & 0.776 & 0.847 &  \textbf{0.794}   & 0.726 & \textbf{0.783} \\ \bottomrule
    \end{tabular}
\end{adjustbox}
    \caption{Cosine similarity of sentence embeddings between the augmented sentence and the original one. IMDB (80) means the number of training samples after sub-sampling is 80. Here $n_\text{aug}=16$ and number of iterations $T$ is 200.}
    \label{tb:semantic}
\end{table*}

\begin{figure*}[t!]
    \centering
    \includegraphics[width=0.9\textwidth]{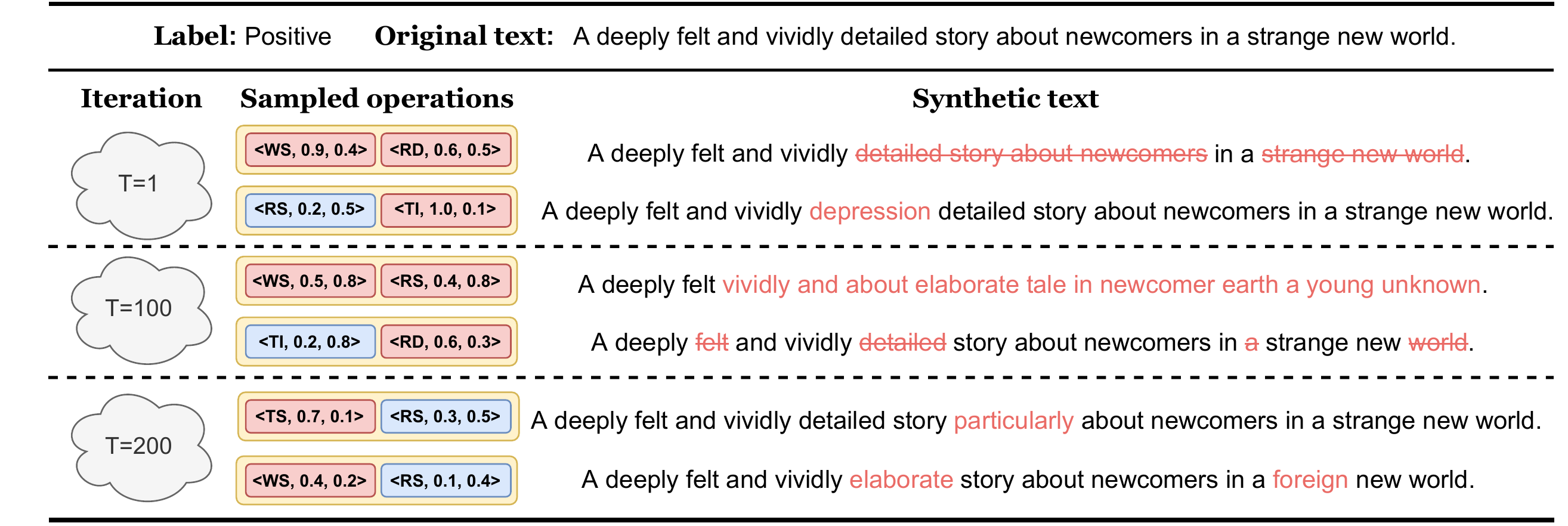}
    \caption{Examples of augmented texts generated by TAA on SST-5 after 1/100/200 iteration of exploration. {\color[HTML]{EA6B66} Red operation} $\mathcal{O}=\left< t,p,\lambda\right>$ indicates it is finally applied according to probability $p$. We use {\bf RS} (Random Swap), {\bf RD} (Random Delete), {\bf TI} (TF-IDF Insert), {\bf TS} (TF-IDF Substitute) and {\bf WS} (WordNet Substitute) for abbreviation.}
    \label{Fig:case study}
\end{figure*}

\subsection{Diversity of Augmented Data}
We evaluate the diversity of the augmented data by computing the Dist-2~\citep{distinct-n}, which measures the number of distinct bi-grams in generated sentences. 
The metric is scaled using the total number of bi-grams in the generated sentence, which ranges from 0 to 1 where a larger value indicates higher diversity.  
As shown in Table~\ref{tb:distinct-n}, TAA achieves the best overall performance, even slightly outperforms the generation-based method, i.e., back-translation. 
The results validate the effectiveness of applying various editing operations in the compositional augmentation policy, and demonstrate the superiority of our well-designed algorithm for automatic hyper-parameters tuning.
% effectiveness of our proposal.

\subsection{Semantics Preservation of Augmented Data}
% \section{Augmentation Policy after Optimization}
The augmented sentence are supposed to preserve the semantic meaning of the original sentence, for enriching rather than deviating from the original data set support. 
We propose to evaluate semantic similarity between the generated sentences and the original ones, based on sentence embedding cosine similarity.
In more detail, for the sentence pair $(x, x_\text{aug})$ consists of the original sentence $x$ and the corresponding augmented text $x_\text{aug}$,
we utilize Sentence-BERT~\citep{reimers2019sentencebert} library, which achieves the state-of-the-art performance on various semantic textual similarity benchmarks, to obtain dense vector representations of sentences $(\mathbf{x}, \mathbf{x_\text{aug}})$. The semantic preservation score $\text{SP} (x, x_\text{aug})$ is defined as:
\begin{equation}
    \text{SP}  (x, x_\text{aug} ) = \frac{\mathbf{x} \cdot \mathbf{x_\text{aug}} } {\| \mathbf{x} \| \|\mathbf{x_\text{aug}} \| }.
\end{equation}
We compute the average semantic preservation score for the whole augmented datasets using different augmentation methods, and the results are listed in Table~\ref{tb:semantic}.
It can be found that our TAA achieves the best results, comparable with the generation-based method back-translation.
We note that our method achieves the highest semantic preservation score when the number of training samples is relatively small, which demonstrates that our method can generalize to extreme low-resource scenarios.

\subsection{Case Study}
\label{subsec:Case Study}
Figure~\ref{Fig:case study} shows some cases of augmented texts generated by TAA. 
Given the original text with the label ``positive", the modifications applied by TAA distort the semantics and sentence structure at the beginning. 
As the number of iteration $T$ increases, TAA captures the feature of the dataset adaptively, helping it achieve a better balance between diversity and quality on augmented samples. 
% Further analysis on the semantic preservation of the augmented texts can be found in Appendix G.

\section{Related Work}
\label{sec:Relate}
Previous data augmentation algorithms in NLP can be categorized into generation-based and editing-based methods. 
For generation-based methods, back-translation~\cite{improving-MT, enhancement-MT, da4sa, pda} generates the paraphrase of a text by translating it to an intermediate language and back. 
% from language A to language B, then back to language A. 
\citet{contextual-augmentation} masks some words then uses a label-conditional language model to predict them. 
\citet{not-enough-data} and \citet{ptm4da} leverage various language models such as GPT-2~\cite{gpt-2} and BART~\cite{BART} to generate a continuation of the original sentences. 
For editing-based methods, \citet{eda} propose the Easy Data Augmentation (EDA) technique, which randomly selects editing operations from four candidates for the augmentation. 
\citet{uda} propose to substitute uninformative words with low TF-IDF scores. %while keeping others unchanged. 
These methods require a lot of prior knowledge to preset their parameters and are prone to fall into the sub-optimum.

% Previous works on data augmentation require a lot of prior knowledge about downstream tasks. 
Recently,  % To lighten the burden of artificial designing and parameters turning, 
some algorithms are proposed for automatically learning augmentation policy in the field of Computer Vision~\cite{autoaugment, PBA, fast-autoaugment, Faster-AutoAugment, dada} and NLP~\cite{dm4dialogue, Rotom}. 
However, their modelings are different from ours. 
Specifically, 
AutoAugment~\cite{autoaugment} establishes a Reinforcement Learning (RL) framework to search for the best augmentation policy.
In order to reduce the time of policy exploration, \citet{PBA} replace the RL framework with a population-based algorithm.
\citet{fast-autoaugment} leverage 5-fold cross-validation and augments the validation set instead of the training set. 
While in our proposed approach, the objective function for policy optimization is designed as the loss that models achieve on the validation set after trained on the augmented set. Besides, we utilize SMBO as the optimization algorithm, which can return a promising result efficiently. 
In the field of NLP, TAA surpasses both EDA~\cite{eda} and LDM~\cite{learning-data-manipulation} via a learnable and compositional augmentation policy.% learning. 

\section{Conclusion}
In this paper, we propose an effective method called Text AutoAugment (TAA) to establish a compositional and learnable paradigm for data augmentation. TAA regards a combination of various editing operations as an augmentation policy and utilizes SMBO for policy learning. Experiments show that TAA can substantially improve the generalization ability of models as well as lighten the burden of artificial augmentation designing.

\section*{Acknowledgements}
We thank all the anonymous reviewers for their constructive comments, and Xuancheng Ren and Guangxiang Zhao for their valuable suggestions in preparing the manuscript. 
This is a joint work between Pattern Recognition Center, WeChat AI, Tencent Inc. and Peking University, and is partly supported by Beijing Academy of Artificial Intelligence (BAAI). 
Xu Sun is the corresponding author of this paper. 

\bibliography{anthology,ori_custom}
\bibliographystyle{acl_natbib}

% \appendix
\newpage
\appendix

\section{Editing Operations for Augmentation}
\label{appsec:editing operations}
\begin{table*}[t!]
\centering
\begin{tabular}{ll}
\toprule
Operation Name        & Description  \\ \midrule
Random Swap (RS)      &  Swap two adjacent words randomly.     \\
Random Delete (RD)   & Delete words randomly. \\
WordNet Substitute (WS)   & Substitute words with their synonyms according to WordNet. \\
TF-IDF Substitute (TS) &  Substitute uninformative words with low TF-IDF scores.  \\
TF-IDF Insert (TI)  &  Insert informative words with high TF-IDF. \\ \bottomrule
\end{tabular}
\caption{Description of all the augment operations in the search space.}
\label{tb:operation-description}
\end{table*}
We use five simple and effect operations including Random Swap~\cite{eda}, Random Delete~\cite{eda}, WordNet Substitute~\cite{YELP, Siamese-Recurrent-Architectures, eda}, TF-IDF Substitute~\cite{uda} and TF-IDF Insert~\cite{uda} as the basic component of our augmentation policy. 
All operations used here are word-level, considering their low complexity and high effectiveness. 
The detailed description is shown in Table~\ref{tb:operation-description}.

\section{Sequential Model-based Global Optimization (SMBO)}
\label{appsec:SMBO}
% Recall our optimization procedure in Section 2.3, 
Recall the TAA framework in Section 2, our objective is to search for the optimal augmentation policy $\mathcal{P}$ that minimizes the following loss: 
\begin{equation}
    \mathcal{J} = \mathcal{J}\left( f, \mathcal{D}_\text{aug}(\mathcal{P}), \mathcal{D}_\text{val} \right).
\label{equ:total loss}
\end{equation}
We leverage the Sequential Model-based Global Optimization (SMBO) as our optimizer for the policy learning.
The optimization procedure is carried out in an iterative manner. 
To build the relation between the policy $\mathcal{P}$ and the objective loss $\mathcal{J}$ and sample the most promising policy at each iteration, 
we use the Tree-structured Parzen Estimator (TPE)~\cite{SMBO} as a surrogate agent $M$ to model the conditional probability $p_M(\mathcal{J} \mid \mathcal{P})$. %between the policy $\mathcal{P}$ and the objective loss $\mathcal{J}$. 
Besides, we employ the following Expected Improvement (EI) criterion as an acquisition function for policy sampling in the current iteration: 
\begin{equation}
\begin{split}
    % \mathrm{EI}_{\mathcal{J}^{\dagger}}(\mathcal{P})&=\mathbb{E} \max \left(\left(\mathcal{J}^{\dagger}-\mathcal{J}\right), 0\right) \\
    \mathrm{EI}(\mathcal{P}) &= \mathbb{E} \left[ \max \left( \mathcal{J}^{\dagger} - \mathcal{J}, 0 \right)\right] \\
    &=\int^{\mathcal{J}^{\dagger}}_{-\infty}\left( \mathcal{J}^{\dagger} - \mathcal{J} \right) p_{M}(\mathcal{J} \mid \mathcal{P}) d \mathcal{J}
\end{split}
\label{equ:EI}
\end{equation}
Here, $\mathcal{J}^{\dagger}$ is a threshold and Eq.~\ref{equ:EI} stands for the expectation under the surrogate model that the loss $\mathcal{J}$ of a policy will exceed (negatively) the threshold $\mathcal{J}^{\dagger}$. 
We expect the loss of the sampled policy in each iteration to be smaller than the current threshold, thus the policy that maximizes the Expected Improvement will be chosen in the next iteration. 
% The greater the reduction, the greater the expected improvement. Under the surrogate agent $p_{M}(\mathcal{J} \mid \mathcal{P})$, the policy that maximizes the expected improvement will be sampled. 

\begin{table*}[htp]
    \centering
    % \small 
    \begin{adjustbox}{max width=0.9\textwidth}
    \begin{tabular}{lcrrrc}
        \toprule
        Dataset & \# Classes & \# Train samples & \# Test samples &  Average length     & Task                \\ \midrule
        IMDB    & 2         & 25,000          & 25,000         & 326      & Sentiment Analysis       \\
        SST-2   & 2         & 7,791           & 1,821          & 18     & Sentiment Analysis         \\
        SST-5   & 5         & 9,643           & 2,210          & 19    & Sentiment Analysis          \\
        TREC    & 6         & 5,452           & 500            & 10        & Question Classification     \\
        YELP-2   & 2         & 560,000         & 38,000         & 139  & Review Classification        \\
        YELP-5   & 5         & 650,000         & 50,000         & 141  & Review Classification          \\ \bottomrule
    \end{tabular}
    \end{adjustbox}
    \caption{Statistics on the datasets.}
    \label{tb:dataset detail}
\end{table*}
% \usepackage{booktabs}
\begin{table*}[t!]
\centering
% \small 
\begin{adjustbox}{max width=0.9\textwidth}
\begin{tabular}{cccccc}
\toprule
Policy & IMDB (80)                               & SST-5 (200)                             & TREC (120)                              & YELP-2 (80)                             & YELP-5 (200)                            \\ \midrule
Op1    & $\left<\text{\text{TS}}, 0.77, 0.07\right>$ & $\left<\text{RD}, 0.44, 0.22\right>$ & $\left<\text{WS}, 0.75, 0.44\right>$ & $\left<\text{TS}, 0.95, 0.26\right>$ & $\left<\text{RD}, 0.34, 0.32\right>$ \\
Op2    & $\left<\text{TS}, 0.50, 0.31\right>$ & $\left<\text{WS}, 0.59, 0.50\right>$ & $\left<\text{WS}, 0.33, 0.26\right>$ & $\left<\text{RD}, 0.08, 0.49\right>$ & $\left<\text{WS}, 0.69, 0.19\right>$ \\
Op3    & $\left<\text{RD}, 0.72, 0.05\right>$ & $\left<\text{TI}, 0.66, 0.11\right>$ & $\left<\text{RS}, 0.60, 0.02\right>$ & $\left<\text{WS}, 0.57, 0.41\right>$ & $\left<\text{RD}, 0.70, 0.15\right>$ \\
Op4    & $\left<\text{TI}, 0.66, 0.09\right>$ & $\left<\text{TS}, 0.11, 0.25\right>$ & $\left<\text{WS}, 0.59, 0.36\right>$ & $\left<\text{RD}, 0.85, 0.02\right>$ & $\left<\text{RD}, 0.85, 0.27\right>$ \\
Op5    & $\left<\text{TI}, 0.69, 0.13\right>$ & $\left<\text{WS}, 0.69, 0.09\right>$ & $\left<\text{TI}, 0.59, 0.17\right>$ & $\left<\text{RS}, 0.30, 0.15\right>$ & $\left<\text{TS}, 0.72, 0.35\right>$ \\
Op6    & $\left<\text{RS}, 0.26, 0.05\right>$ & $\left<\text{WS}, 0.41, 0.07\right>$ & $\left<\text{WS}, 0.63, 0.34\right>$ & $\left<\text{TS}, 0.55, 0.05\right>$ & $\left<\text{RD}, 0.58, 0.47\right>$ \\
Op7    & $\left<\text{TS}, 0.77, 0.50\right>$ & $\left<\text{TS}, 0.69, 0.02\right>$ & $\left<\text{TS}, 0.22, 0.37\right>$ & $\left<\text{TS}, 0.63, 0.34\right>$ & $\left<\text{RS}, 0.48, 0.22\right>$ \\
Op8    & $\left<\text{TS}, 0.36, 0.25\right>$ & $\left<\text{WS}, 0.99, 0.22\right>$ & $\left<\text{RS}, 0.96, 0.03\right>$ & $\left<\text{WS}, 0.13, 0.28\right>$ & $\left<\text{TS}, 0.72, 0.50\right>$ \\ \bottomrule
\end{tabular}
\end{adjustbox}
\caption{TAA policies searched in the low-resource regime. Each policy finally consists of $8$ atomic editing operations and each operation satisfies the form of $\mathcal{O}=\left<t,p,\lambda\right>$. We use {\bf RS} (Random Swap), {\bf RD} (Random Delete),  {\bf WS} (WordNet Substitute), {\bf TI} (TF-IDF Insert), {\bf TS} (TF-IDF Substitute) for simplification.}
\label{tb:searched-policy}
\end{table*}

Instead of directly representing $p_M(\mathcal{J} \mid \mathcal{P})$ for calculating the Expected Improvement, the TPE builds a model of $p_M(\mathcal{P} \mid \mathcal{J})$ by applying Bayes rule. 
The TPE splits the historical observations of policies in two groups: the best performing one (e.g., the upper quartile) and the rest. 
The threshold $\mathcal{J}^{\dagger}$ is defined as the splitting value for the two groups and the likelihood probability for being in each of these groups is modeled as:
\begin{equation}
    p_M(\mathcal{P} \mid \mathcal{J})=\left\{\begin{array}{ll}l(\mathcal{P}) & \text { if } \mathcal{J}<\mathcal{J}^{\dagger} \\ g(\mathcal{P}) & \text { if } \mathcal{J}>\mathcal{J}^{\dagger}\end{array}\right.
\end{equation}
% We employ variable kernel density estimation [33] on graph-structured search space S to approximate the criterion. 
Here, $l(\mathcal{P})$ models the distribution of previous sampled policies whose loss is less than the threshold, and $g(\mathcal{P})$ models the distribution of the other policies whose loss is greater than the threshold. 
The two densities $l$ and $g$ are modeled using Parzen estimators (also known as kernel density estimators), which are a simple average of kernels centered on existing data points. 
With Bayes Rule, we can prove that the Expected Improvement which we are trying to maximize is proportional to the ratio $l(\mathcal{P}) / g(\mathcal{P})$:
\begin{equation}
\small
    \begin{aligned} \mathrm{EI}(\mathcal{P}) &=\int_{-\infty}^{\mathcal{J}^{\dagger}}\left(\mathcal{J}^{\dagger}-\mathcal{J}\right) p_{M}(\mathcal{J} \mid \mathcal{P}) d \mathcal{J} \\ &=\int_{-\infty}^{\mathcal{J}^{\dagger}}\left(\mathcal{J}^{\dagger}-\mathcal{J}\right) \frac{p_M(\mathcal{P} \mid \mathcal{J}) p(\mathcal{J})}{p(\mathcal{P})} d \mathcal{J} \\ &=\frac{l(\mathcal{P})}{p(\mathcal{P})} \int_{-\infty}^{\mathcal{J}^{\dagger}}\left(\mathcal{J}^{\dagger}-\mathcal{J}\right) p(\mathcal{J}) d \mathcal{J} \\ &=\frac{l(\mathcal{P})}{p(\mathcal{P})}\left[\gamma \mathcal{J}^{\dagger}-\int_{-\infty}^{\mathcal{J}^{\dagger}} p(\mathcal{J}) d \mathcal{J}\right] \\ &=\frac{l(\mathcal{P})}{\gamma l(\mathcal{P})+(1-\gamma) g(\mathcal{P})}\left[\gamma \mathcal{J}^{\dagger}-\int_{-\infty}^{\mathcal{J}^{\dagger}} p(\mathcal{J}) d \mathcal{J}\right] \\ & \propto\left(\gamma+\frac{g(\mathcal{P})}{l(\mathcal{P})}(1-\gamma)\right)^{-1} \end{aligned}
\end{equation}
Accordingly, the maximization of the Expected Improvement can be achieved by maximizing the ratio $l(\mathcal{P}) / g(\mathcal{P})$. 
In other words, we should sample the polices which are more likely under $l(\mathcal{P})$ than under $g(\mathcal{P})$. 
The TPE works by drawing sample polices from $l(\mathcal{P})$, evaluating them in terms of $l(\mathcal{P}) / g(\mathcal{P})$, and returning the set that yields the highest value under $l(\mathcal{P}) / g(\mathcal{P})$ corresponding to the greatest expected improvement. 
These policies are then evaluated on the objective function and the results are merged into the observation history. 
The algorithm builds $l(\mathcal{P})$ and $g(\mathcal{P})$ using the history to update the probability model $M$ of the objective function that improves with each iteration.

\section{Statistics of Datasets}
\label{appsec:datasets statistics}
We conduct experiments on six popular datasets including IMDB~\cite{IMDB}, SST-2, SST-5~\cite{SST}, TREC~\cite{TREC}, YELP-2 and YELP-5~\cite{YELP}. The statistics of datasets used are listed in Table~\ref{tb:dataset detail}.

\section{Searched Policy}
\label{appsec:searched policy}
The policies searched by our TAA algorithm are listed in Table~\ref{tb:searched-policy}. 
Each policy finally consists of $8$ atomic editing operations and each operation satisfies the form of $\mathcal{O}=\left<t,p,\lambda\right>$.
All the policies can be used directly to augment the full training set to further boost the model performance on the corresponding downstream tasks.

\section{Ablation Study of TAA Policy}
\label{appsec:ablation study}
\label{subsec:Ablation Study}

\begin{table*}[t!]
\centering
% \small 
\begin{adjustbox}{max width=0.95\textwidth}
    % \begin{tabular}{lccccccc}
    %     \toprule
    %     \multirow{2}{*}{Method} & \multicolumn{3}{c}{Low-resource Regime} & \multicolumn{3}{c}{Class-imbalanced Regime}     & Overall                                                                                                       \\ \cmidrule(lr){2-4} \cmidrule(lr){5-7} \cmidrule(lr){8-8}
    %                               & \begin{tabular}[c]{@{}c@{}}IMDB\\ (80)\end{tabular}          & \begin{tabular}[c]{@{}c@{}}SST-5\\ (200)\end{tabular}                 & \begin{tabular}[c]{@{}c@{}}TREC\\ (120)\end{tabular} & \begin{tabular}[c]{@{}c@{}}IMDB\\ (20:1000)\end{tabular} & \begin{tabular}[c]{@{}c@{}}IMDB\\ (50:1000)\end{tabular} & \begin{tabular}[c]{@{}c@{}}IMDB\\ (100:1000)\end{tabular} & Average \\ \midrule
    %     TAA-R                      & 72.64\small$\pm$5.55                      & 38.60\small$\pm$2.12                             & 77.72\small$\pm$4.01             &         55.72\small$\pm$3.74          &   63.60\small$\pm$5.71        &     74.37\small$\pm$4.22     & 63.78\small$\pm$4.23  \\
    %     \textbf{TAA}               & \textbf{73.54}\small$\pm$2.72             & \textbf{39.46}\small$\pm$2.25                    & \textbf{79.99}\small$\pm$3.68    &       \textbf{56.91}\small$\pm$2.80       &    \textbf{66.41}\small$\pm$4.85       &   \textbf{77.87}\small$\pm$3.18   & \textbf{65.70}\small$\pm$3.25       \\ \bottomrule
    % \end{tabular}
    \begin{tabular}{lccccccc}
        \toprule
        \multirow{2}{*}{Method} & \multicolumn{3}{c}{Low-resource Regime} & \multicolumn{3}{c}{Class-imbalanced Regime}     & Overall                                                                                                       \\ \cmidrule(lr){2-4} \cmidrule(lr){5-7} \cmidrule(lr){8-8}
                                   & \begin{tabular}[c]{@{}c@{}}IMDB\\ (80)\end{tabular}          & \begin{tabular}[c]{@{}c@{}}SST-5\\ (200)\end{tabular}                 & \begin{tabular}[c]{@{}c@{}}TREC\\ (120)\end{tabular} & \begin{tabular}[c]{@{}c@{}}IMDB\\ (20:1000)\end{tabular} & \begin{tabular}[c]{@{}c@{}}IMDB\\ (50:1000)\end{tabular} & \begin{tabular}[c]{@{}c@{}}IMDB\\ (100:1000)\end{tabular} & Average \\ \midrule
        TAA-R                      & 72.64\small$\pm$5.55                      & 38.60\small$\pm$2.12                             & 77.72\small$\pm$4.01             &         55.72\small$\pm$3.74          &   63.60\small$\pm$5.71        &     74.37\small$\pm$4.22     & 63.78\small$\pm$4.23  \\
        \textbf{TAA}               & \textbf{75.68}$^*$\small$\pm$3.27             & \textbf{40.28}$^{**}$\small$\pm$1.80                    & \textbf{81.47}$^{**}$\small$\pm$3.87    &       \textbf{56.92}\small$\pm$2.80       &    \textbf{66.73}$^*$\small$\pm$4.20       &   \textbf{77.87}$^*$\small$\pm$3.18   & \textbf{66.49}\small$\pm$3.19       \\ \bottomrule
    \end{tabular}
    \end{adjustbox}
    \caption{Test accuracy (\%) for TAA-R and TAA in low-resource and class-imbalanced regime. $^*$ and $^{**}$ indicate statistically significant (p < .05 and p < .01) improvements over TAA-R.}
    \label{tb:TAA-R}
\end{table*}

\subsection{Searching Algorithm Ablation}
To verify the effectiveness of our optimization algorithm, we re-execute the experiment on the same search space described in Section 2.1, but do NOT conduct any policy optimization.
For each text in the training set, a random policy is sampled to synthesize the augmented data. We call this method Text AutoAugment-Random (\textbf{TAA-R}). 
Table~\ref{tb:TAA-R} illustrates the results of TAA and TAA-R. Generally, the augmentation policy after optimization performs better than the random policy by 1.92\%. 
% Surprisingly, the randomly sampled policy also performs well. Similar results are also found in the AutoAugment~\cite{autoaugment}. One possible explanation is that 
Note that we incorporate prior knowledge in Section 2.1 to constrain the range of the operation magnitude, which ensures the performance of the operation and avoids generating bad samples.

\subsection{Operation Type Ablation}
%\begin{table*}[t]
%\centering
%\scalebox{0.9}{
%\begin{tabular}{@{}lllllll@{}}
%\toprule
%Dataset        & w/o RS         & w/o RD         & w/o TI         & w/o TS         & w/s WS         & Full Space     \\ \midrule
%IMDB (80)       & 73.90\small$\pm$2.21 & 72.99\small$\pm$3.44 & 74.91\small$\pm$3.06 & 73.95\small$\pm$3.93 & 72.55\small$\pm$5.22 & 74.02\small$\pm$3.18 \\
%SST-2 (50:1000) & 63.58\small$\pm$6.76 & 64.03\small$\pm$5.01 & 64.73\small$\pm$6.04 & 65.20\small$\pm$4.95 & 66.94\small$\pm$4.48 & 65.67\small$\pm$6.10 \\ \bottomrule
%\end{tabular}}
%\caption{Search space ablation test (\%). We use {\bf RS} (Random Swap), {\bf RD} (Random Delete), {\bf TI} (TF-IDF Insert), {\bf TS} (TF-IDF Substitute) for simplification.}
%\end{table*}
\begin{table*}[t!]
\centering
\begin{tabular}{lcccccc}
\toprule
Dataset        & Full Space & w/o RS & w/o RD & w/o TI & w/o TS & w/o WS \\ \midrule
IMDB(80)       & 75.68      & $-$0.12  & $-$1.03  & $+$0.89  & $-$0.07  & $-$1.47  \\
SST-2(50:1000) & 66.05      & $-$2.09  & $-$1.64  & $-$0.94  & $-$0.47  & $+$1.27  \\ \bottomrule
\end{tabular}
\caption{Test accuracy (\%) in the ablation study of operation type. We use {\bf RS} (Random Swap), {\bf RD} (Random Delete),  {\bf WS} (WordNet Substitute), {\bf TI} (TF-IDF Insert), {\bf TS} (TF-IDF Substitute) for simplification.}
\label{tb:search-space-ablation}
\end{table*}
We conduct operation type ablation study to examine the effect of the operation types search space. 
Specifically, we eliminate one operation type while keeps others for searching the optimal policy, and evaluate the task performance using the learned policy.
The task performance difference with different ablated operations on IMDB and SST-2 dataset are shown in Table~\ref{tb:search-space-ablation}.
We find that the elimination of operation type generally leads to a decrease of task performance, indicating that the effect of different operations are complementary with each other. 
We attribute it to that more operations types will improve the diversity of the combinations of text manipulation, thus boosting the dataset quality and benefiting the generalizability of the model.

\end{document}